\documentclass[10pt,twocolumn,letterpaper]{article}

\usepackage{cvpr}              

\usepackage{graphicx}
\usepackage{amsmath}
\usepackage{amssymb}
\usepackage{booktabs}
\usepackage{amsfonts}
\usepackage{multirow}

\usepackage[accsupp]{axessibility}  

\usepackage[pagebackref,breaklinks,colorlinks]{hyperref}
\usepackage{pifont}

\usepackage[capitalize]{cleveref}
\crefname{section}{Sec.}{Secs.}
\Crefname{section}{Section}{Sections}
\Crefname{table}{Table}{Tables}
\crefname{table}{Tab.}{Tabs.}

\def\cvprPaperID{3664} 
\def\confName{CVPR}
\def\confYear{2022}

\newlength\savewidth\newcommand\shline{\noalign{\global\savewidth\arrayrulewidth
		\global\arrayrulewidth 1pt}\hline\noalign{\global\arrayrulewidth\savewidth}}
\newcommand{\tablestyle}[2]{\setlength{\tabcolsep}{#1}\renewcommand{\arraystretch}{#2}\centering\footnotesize}
\newcommand{\cmark}{\ding{51}}%
\newcommand{\tabincell}[2]{\begin{tabular}{@{}#1@{}}#2\end{tabular}}

\begin{document}

\title{Towards Robust Vision Transformer}

\author{Xiaofeng Mao\textsuperscript{1}
\quad
Gege Qi\textsuperscript{1}
\quad
Yuefeng Chen\textsuperscript{1} 
\quad
Xiaodan Li\textsuperscript{1} 
\quad
Ranjie Duan\textsuperscript{2}
\quad
Shaokai Ye\textsuperscript{3} \\
\quad
Yuan He\textsuperscript{1} 
\quad
Hui Xue\textsuperscript{1} \\
\textsuperscript{1}Alibaba Group
\quad
\textsuperscript{2}Swinburne University of Technology
\quad
\textsuperscript{3}EPFL \\
{\tt\small\{mxf164419,qigege.qgg,yuefeng.chenyf,fiona.lxd\}@alibaba-inc.com}
}

\maketitle

\begin{abstract}
Recent advances on Vision Transformer (ViT) and its improved variants have shown that self-attention-based networks surpass traditional Convolutional Neural Networks (CNNs) in most vision tasks. However, existing ViTs focus on the standard accuracy and computation cost, lacking the investigation of the intrinsic influence on model robustness and generalization. In this work, we conduct systematic evaluation on components of ViTs in terms of their impact on robustness to adversarial examples, common corruptions and distribution shifts. We find some components can be harmful to robustness. By leveraging robust components as building blocks of ViTs, we propose \textbf{Robust Vision Transformer (RVT)}, which is a new vision transformer and has superior performance with strong robustness. Inspired by the findings during the evaluation, we further propose two new plug-and-play techniques called position-aware attention scaling and patch-wise augmentation to augment our RVT, which we abbreviate as RVT$^{*}$. The experimental results of RVT on ImageNet and six robustness benchmarks demonstrate its advanced robustness and generalization ability compared with previous ViTs and state-of-the-art CNNs. Furthermore, RVT-S$^{*}$ achieves Top-1 rank on multiple robustness leaderboards including ImageNet-C, ImageNet-Sketch and ImageNet-R.
\end{abstract}


\section{Introduction}

Following the popularity of transformers in Natural Language Processing (NLP) applications, e.g., BERT~\cite{DBLP:conf/naacl/DevlinCLT19} and GPT~\cite{radford2018improving}, there has sparked particular interest in investigating whether transformer can be a primary backbone for computer vision applications previously dominated by Convolutional Neural Networks (CNNs). Recently, Vision Transformer (ViT)~\cite{dosovitskiy2020image} successfully applies a pure transformer for classification which achieves an impressive speed-accuracy trade-off by capturing long-range dependencies via self-attention. Base on this seminal work, numerous variants have been proposed to improve ViTs from different perspectives containing training data efficiency~\cite{touvron2020training}, self-attention mechanism~\cite{liu2021swin}, introducing convolution~\cite{li2021localvit, wu2021cvt,yuan2021incorporating} or pooling layers~\cite{wang2021pyramid,heo2021rethinking}, etc. However, these works only focus on the standard accuracy and computation cost, lacking the investigation of the intrinsic influence on model robustness and generalization.

\begin{figure}
\vspace{-5pt}
    \centering
    \includegraphics[width=1.\linewidth]{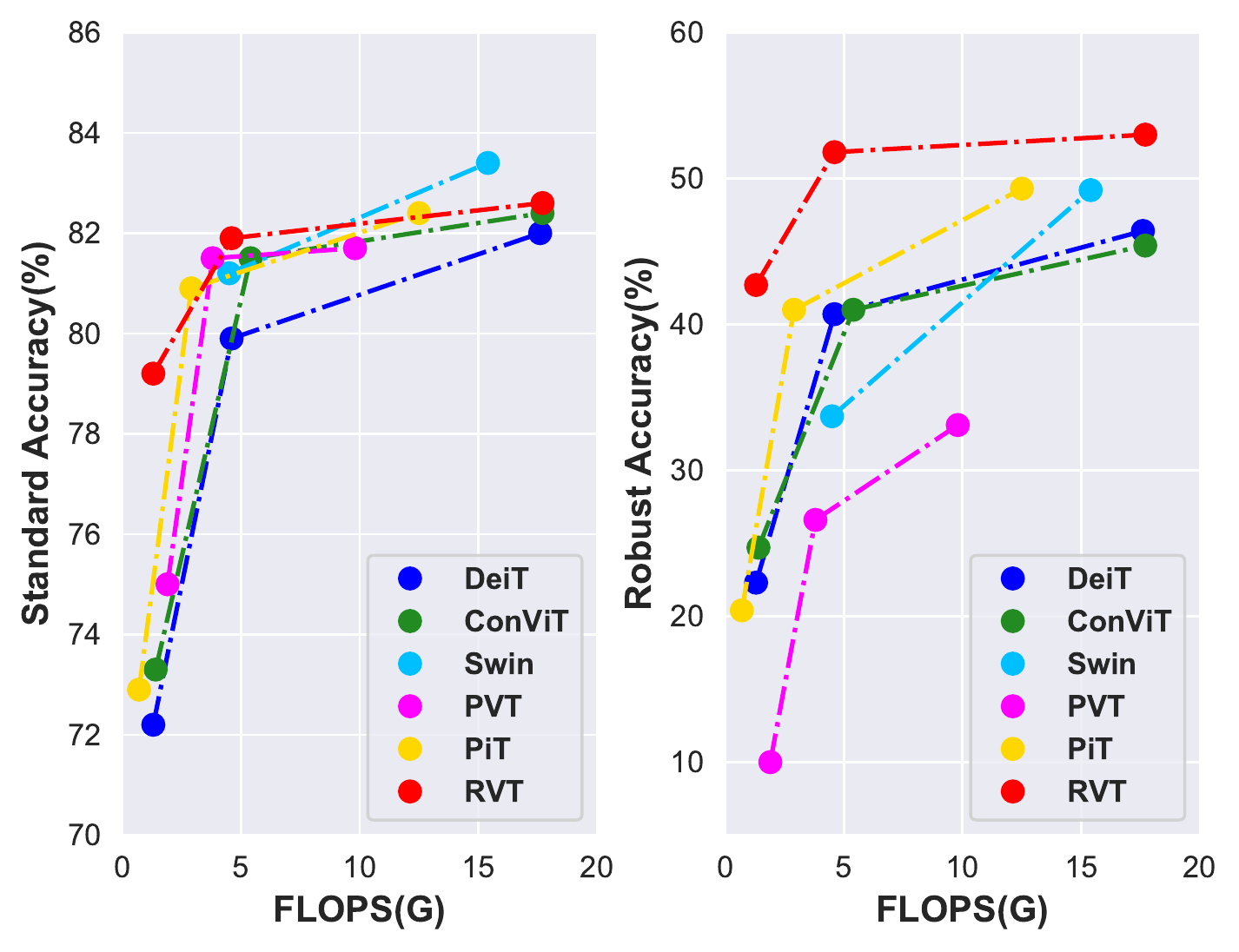}
    \caption{Comparison between RVT and the baseline transformers. The robust accuracy in figure is recorded under FGSM~\cite{goodfellow2014explaining} adversary.}
    \label{fig:teaser1}
    \vspace{-1em}
\end{figure}

In this work, we take initiatives to explore a ViT model with strong robustness. To this end, we first give an empirical assessment of existing ViT models in Figure~\ref{fig:teaser1}. Surprisingly, although all ViT variants reproduce the standard accuracy claimed in the paper, some of their modifications may bring devastating damages on the model robustness. A vivid example is PVT~\cite{wang2021pyramid}, which achieves a high standard accuracy but suffered with large drop of robust accuracy. 
We show that PVT-Small obtains only 26.6\% robust accuracy, which is 14.1\% lower than original DeiT-S in Figure~\ref{fig:teaser1}.

To demystify the trade-offs between accuracy and robustness, we analyze ViT models with different patch embedding, position embedding, transformer blocks and classification head whose impact on the robustness that has never been thoroughly studied. Based on the valuable findings revealed by exploratory experiments, we propose a Robust Vision Transformer (RVT), which has significant improvement on robustness, but also exceeds most other transformers in accuracy. In addition, we propose two new plug-and-play techniques to further boost the RVT. The first is Position-Aware Attention Scaling (PAAS), which plays the role of position encoding in RVT. PAAS improves the self-attention mechanism by filtering out redundant and noisy position correlation and activating only major attention with strong correlation, which leads to the enhancement of model robustness. The second is a simple and general patch-wise augmentation method for patch sequences which adds rich affinity and diversity to training data. Patch-wise augmentation also contributes to the model generalization by reducing the risk of over-fitting. With the above proposed methods, we can build an augmented Robust Vision Transformer$^{*}$ (RVT$^{*}$). Contributions of this paper are three-fold:

\begin{itemize}
\item We give a systematic robustness analysis of ViTs and reveal harmful components. Inspired by it, we reform robust components as building blocks as a new transformer, named Robust Vision Transformer (RVT).

\item To further improve the RVT, we propose two new plug-and-play techniques called position-aware attention scaling and patch-wise augmentation. Both of them can be applied to other ViT models and yield significant enhancement on robustness and standard accuracy.

\item Experimental results on ImageNet and six robustness benchmarks show that RVT exhibits best trade-offs between standard accuracy and robustness compared with previous ViTs and CNNs. Specifically, RVT-S$^{*}$ achieves Top-1 rank on ImageNet-C, ImageNet-Sketch and ImageNet-R.
\end{itemize}

\section{Related Work}
\label{sec:2}
\textbf{Robustness Benchmarks.} The rigorous benchmarks are important for evaluating and understanding the robustness of deep models. Early works focus on the model safety under the adversarial examples with constrained perturbations~\cite{goodfellow2014explaining, szegedy2013intriguing}. In real-world applications, the phenomenon of image corruption or out-of-distribution is more commonly appeared. 
Driven by this, ImageNet-C~\cite{hendrycks2019benchmarking} benchmarks the model against image corruption which simulates distortions from real-world sources. ImageNet-R~\cite{hendrycks2020many} and ImageNet-Sketch~\cite{wang2019learning} collect the online images consisting of naturally occurring distribution changes such as image style, to measure the generalization ability to new distributions at test time. In this paper, we adopt all the above benchmarks as the fair-minded evaluation metrics.

\textbf{Robustness Study for CNNs.} The robustness research of CNNs has experienced explosive development in recent years. Numerous works conduct thorough study on the robustness of CNNs and aim to strengthen it in different ways, e.g., stronger data augmentation~\cite{hendrycks2020many, hendrycks2019augmix, rusak2020simple}, carefully designed~\cite{su2018robustness, wu2020wider} or searched~\cite{guo2020meets, dong2020adversarially} network architecture, improved training strategy~\cite{madry2017towards, li2020shape, xie2020self}, quantization~\cite{lin2019defensive} and pruning~\cite{ye2019adversarial} of the weights, better pooling~\cite{zhang2019making, vasconcelos2020effective} or activation functions~\cite{xie2020smooth}, etc. Although the methods mentioned above perform well on CNNs, there is no evidence that they also keep the effectiveness on ViTs. A targeted research for improving the robustness of ViTs is still blank. 

\begin{figure*}
    \centering
    \includegraphics[scale=.52]{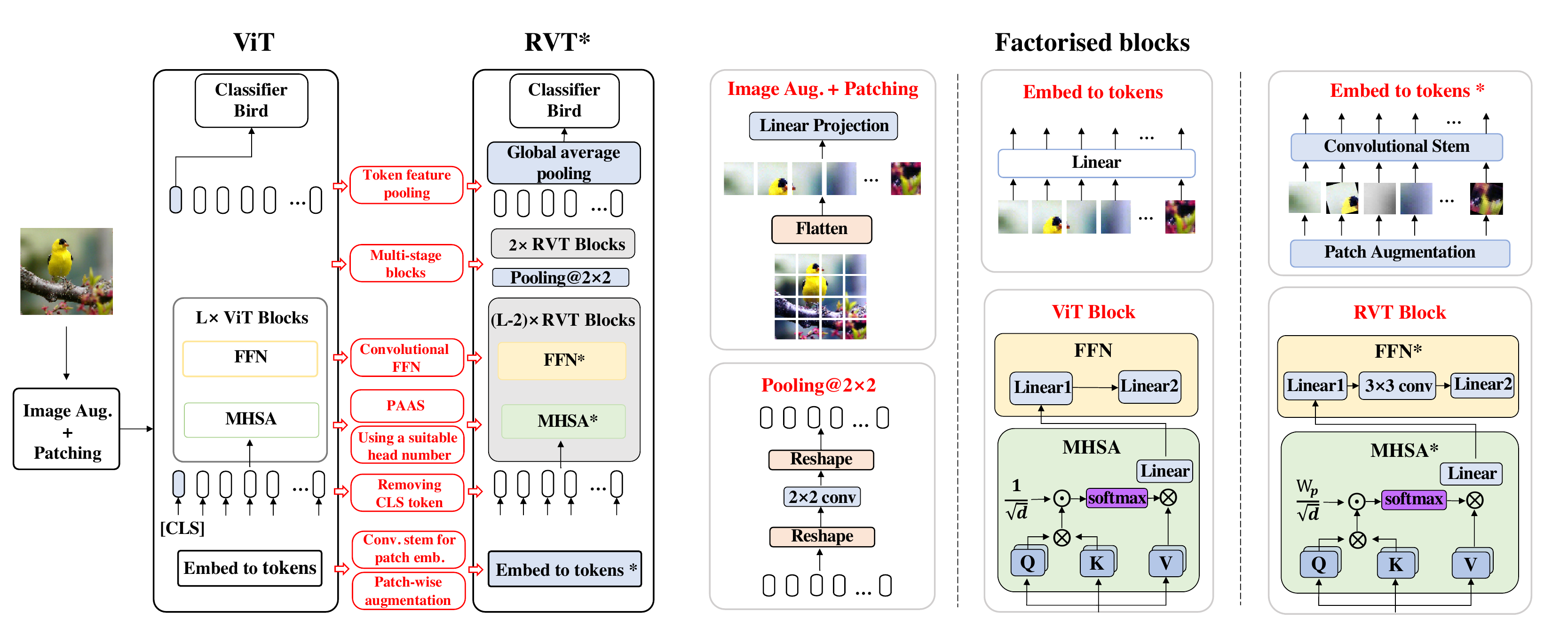}
    \vspace{-4.5mm}
    \caption{\textbf{Overall architecture of the proposed Robust Vision Transformer (RVT).}}
    \label{fig:overall_arch}
    \vspace{-3.5mm}
\end{figure*}

\textbf{Robustness Study for ViTs.} Until now, there are several works attempting at studying the robustness of ViTs. Early works focus on the adversarial robustness of ViTs. They find that ViTs are more adversarially robust than CNNs~\cite{shao2021adversarial} and the transferability of adversarial examples between CNNs and ViTs is remarkably low\cite{mahmood2021robustness}. Follow up works~\cite{bhojanapalli2021understanding,paul2021vision} extend the robustness study on ViTs to much common image corruption and distribution shift, and indicate ViTs are more robust learners. Although some findings are consistent with above works, in this paper, we do not make simple comparison of robustness between ViTs and CNNs, but take a step further by analyzing the detailed robust components in ViT and its variants. Based on the analysis, we design a robust vision transformer and introduce two novel techniques to further reduce the fragility of ViT models.

\section{Robustness Analysis of Designed Components}
\label{sec:3}

We give the robustness analysis of four main components in ViTs: patch embedding, position embedding, transformer blocks and classification head. DeiT-Ti~\cite{touvron2020training} is used as the base model. All the robustness benchmarks mentioned in section~\ref{sec:2} are considered comprehensively. There is a positive correlation between these benchmarks in most cases. Due to the limitation of space, we show the robust accuracy under FGSM~\cite{goodfellow2014explaining} adversary in the main body and other results in Appendix A.

\subsection{Patch Embedding}
\textbf{F1: } \textbf{Low-level feature of patches helps for the robustness.} ViTs~\cite{dosovitskiy2020image} tokenize an image by splitting it into patches with size of 16$\times$16 or 32$\times$32. Such simple tokenization makes the models hard to capture low-level structures such as edges and corners. 
To extract low-level features of patches, CeiT~\cite{yuan2021incorporating}, LeViT~\cite{graham2021levit} and TNT~\cite{han2021transformer} use a convolutional stem instead of the original linear layer, T2T-ViT~\cite{yuan2021tokens} leverages self-attention to model dependencies among neighboring pixels. However, these methods merely focus on the standard accuracy. To answer how is the robustness affected by leveraging low-level features of patches, we compare the original linear projection with two new convolution and tokens-to-tokens embedders, proposed by CeiT and T2T-ViT respectively. As shown in Table \ref{tab:analysis_other}, low-level patch embedding has a positive effect on the model robustness and standard accuracy as more detailed visual features are exploited. Among them tokens-to-tokens embedder is the best, but it has quadratic complexity with the expansion of image size. We adopt the convolutional embedder with less computation cost.

\begin{table}[h!]
	\vspace{-5pt}
	\centering
	\small
	\tablestyle{8pt}{1.1}
	\begin{tabular}{lr|c|c}
		&positional embedding & Acc & Robust Acc  \\
		\shline
		(i) &none   & 68.3 & 15.8 \\ 
		(ii) &learned absolute position & 72.2 & \textbf{22.3}  \\
		(iii)& sin-cos absolute position & 72.0 & 21.9  \\ 
		(iv)&learned relative position~\cite{shaw2018self} & 71.8 & 22.3 \\
		(v)&input-conditioned position~\cite{chu2021we} & \textbf{72.4} & 21.5 \\
	\end{tabular}
	\vspace{.5em}
	\caption{\textbf{Effect of different positional embeddings.} We use Deit-Ti as the base model.  
		\label{tab:analysis_posemb}
	}
	\vspace{-10pt}
\end{table}

\subsection{Position Embedding}
\label{sec:2.2}
\textbf{F2: }\textbf{Position encoding is critical for learning shape-bias based semantic features which are robust to texture changes. Besides, existing position encoding methods have no big impact on the robustness.} We first explore the necessity of position embeddings. Previous work~\cite{chu2021we} shows ViT trained without position embeddings has 4\% drop of standard accuracy. In this work, we find this gap even can be larger on robustness. In Appendix A, we find with no position encoding, ViT fails to recognize shape-bias objects, which leads to 8\% accuracy drop on ImageNet-Sketch. Concerning the ways of positional encoding, learned absolute, sin-cos absolute, learned relative~\cite{shaw2018self}, input-conditioned~\cite{chu2021we} position representations are compared. In Table \ref{tab:analysis_posemb}, the result suggests that most position encoding methods have no big impact on the robustness, and a minority even have a negative effect. Especially, CPE~\cite{chu2021we} encodes position embeddings conditioned on inputs. Such a conditional position representation makes it changed easily with the input, and causes the poor robustness. The fragility of position embeddings also motivates us to design a more robust position encoding method. 

\begin{table}[h!]
    \small
  \centering
        \tablestyle{5pt}{1.05}
        \caption{Ablations on other ViT components, where \cmark indicates the use of the corresponding component. }
        \vspace{-0.7mm}
  \begin{tabular}{c|c|c|c|c|c|c|c}
  \toprule
  \multicolumn{3}{c|}{Patch Emb.} 
  & Local & Conv. & \multirow{2}{*}{$\mathtt{CLS}$} & \multirow{2}{*}{Acc} & Rob. \\
  Linear & Conv. & T2T & SA & FFN & & & Acc \\
  \midrule
  \cmark & &  &  & & \cmark & 72.2 & 22.3 \\
  & \cmark & &  & & \cmark & 73.6 & 23.2\\
  &  & \cmark &  & & \cmark & 74.9 & 25.4\\
  \cmark &  & & \cmark  &  & \cmark & 69.1 & 21.0\\
  \cmark &  & & & \cmark &  & 73.9 & 31.9\\
  \cmark &  & &  &  & \cmark & 72.4 & 28.4 \\
  \bottomrule
  \end{tabular}
  \label{tab:analysis_other}
\end{table}

\subsection{Transformer Blocks} 
\textbf{F3: }\textbf{An elaborate multi-stage design is required for constructing robust vision transformers.} Modern CNNs always start with a feature of large spatial sizes and a small channel size and gradually increase the channel size while decreasing the spatial size. The different sizes of feature maps constitute the multi-stage convolution blocks. As shown by previous works~\cite{cohen2016inductive}, such a design contributes to the expressiveness and generalization performance of the network. PVT~\cite{wang2021pyramid}, PiT~\cite{heo2021rethinking} and Swin~\cite{liu2021swin} employ this design principle into ViTs. To measure the robustness variance with changing of stage distribution, we slightly modify the DeiT-Ti architecture to get five variants (V2-V6) in Table~\ref{tab:analysis_stage}. We keep the overall number of transformer blocks consistent to 12 and replace some of them with smaller or larger spatial resolution. Detailed architecture is shown in Appendix A. By comparing with DeiT-Ti, we find all five variants improve the standard accuracy, benefit from the extraction of hierarchical image features. In terms of robustness, transformer blocks with different spatial sizes show different effects. An experimental conclusion is that the model will get worse on robustness when it contains more transformer blocks with large spatial resolution. On the contrary, reducing the spatial resolution gradually at later transformer blocks contributes to the modest enhancement of robustness. Besides, we also observe that having more blocks with larger input spatial size will increase the number of FLOPs and memory consumption. To achieve the best trade-off on speed and performance, we think V2 is the most compromising choice in this paper. 

\textbf{F4: }\textbf{Robustness can be benefited from the completeness and compactness among attention heads, by choosing an appropriate head number.} ConViT~\cite{d2021convit}, Swin~\cite{liu2021swin}  and LeViT~\cite{graham2021levit} both use more self-attention heads and smaller dimensions of keys and queries to achieve better performance at a controllable FLOPs. To study how does the number of heads affect the robustness, we train DeiT-Ti with different head numbers. Once the number of heads increases, we meanwhile reduce the head dimensions to ensure the overall feature dimensions are unchanged. Similar with generally understanding in NLP~\cite{michel2019sixteen}, we find the completeness and compactness among attention heads are important for ViTs. As shown in the Table~\ref{analysis:head}, the robustness and standard accuracy still gain great improvement with the head increasing till to 8. We think that an appropriate number of heads supplies various aspects of attentive information on the input. Such complete and non-redundant attentive information also introduces more fine-grained representations which are prone to be neglect by model with less heads, thus increases the robustness. 

\begin{table}[h]
	\centering
	\small
		\tablestyle{5pt}{1.05}
		\begin{tabular}{c|c|cc|cc}
			variants &{\scriptsize[{S$_{1}$}, {S$_{2}$}, {S$_{3}$}, {S$_{4}$}]} & FLOPs & Mem & Acc & Robust Acc \\
			\shline
			V1 & [0, 0, 12, 0] & 1.3 & 1.1 & 72.2 & 22.3  \\ 
			\underline{V2} & \underline{[0, 0, 10, 2]} & \textbf{1.2} & \textbf{1.1} & 74.8 & \textbf{24.3} \\ 
			V3 & [0, 2, 10, 0] & 1.5 & 1.7 & 73.8 & 22.0 \\ 
			V4 & [0, 2, 8, 2] & 1.4 & 1.7 & 76.4 &  22.3 \\ 
			V5 & [2, 2, 8, 0] & 3.4 & 6.0 & 73.4 & 17.0 \\ 
			V6 & [2, 2, 6, 2] & 3.4 & 6.0 & \textbf{76.4} & 17.5 \\ 
	\end{tabular}
	\caption{\textbf{Effect of stage  distribution.} We ablate the number of blocks in stages $S_{1}$, $S_{2}$, $S_{3}$, $S_{4}$ of \textbf{DeiT-Ti}, where $S_{1}$ is the stage with the largest 56 $\times$ 56 input spatial dimension, and gradually reduced to half of the original in later stages. The GPU memory consumption is tested on input with batch size of 64. 
	\label{tab:analysis_stage} }
\end{table}

\begin{table}[h]
	\centering
	\small
		\tablestyle{5pt}{1.05}
		\begin{tabular}{c|c|c|c|c|c|c}
		\shline
			Heads & 1 & 2 & 4 & 6 & 8 & 12 \\
			\shline
			Acc & 69.0 & 71.7 & 73.1 & 73.4 & \textbf{73.9} & 73.5  \\ 
			Rob. Acc & 17.6 & 21.4 & 22.8 &  24.6 & \textbf{25.2} & 24.7 \\ 
	\end{tabular}
	\caption{\textbf{The performance variance with the number of heads.} \textbf{DeiT-Ti} with head number of 1, 2, 4, 6, 8 and 12 are trained for comparison. 
	\label{analysis:head} }
		\vspace{-10pt}
\end{table}

\textbf{F5: }\textbf{The locality constraints of self-attention layer may do harm for the robustness. } Vanilla self-attention calculates the pair-wise attention of all sequence elements. But for image classification, local region needs to be paid more attention than remoter regions. Swin~\cite{liu2021swin} limits the self-attention computation to non-overlapping local windows on the input. This hard coded locality of self-attention enjoys great computational efficiency and has linear complexity with respect to image size. Although Swin can also get competitive accuracy, in this work we find such local window self-attention is harmful to the model robustness. The result in Table~\ref{tab:analysis_other} shows after modifying self-attention to the local version, the robust accuracy is getting worse. We think this phenomenon may be partly caused by the destruction of long-range dependencies modeling in ViTs.

\textbf{F6: }\textbf{Feed-forward networks (FFN) can be extended to convolutional FFN by encoding multiple tokens in local regions. Such information exchange of local tokens in FFN makes ViTs more robust. } LocalViT~\cite{li2021localvit} and CeiT~\cite{yuan2021incorporating} introduce connectivity of local regions into ViTs by adding a depth-wise convolution in feed-forward networks (FFN). Our experiment in Table~\ref{tab:analysis_other} verifies that the convolutional FFN greatly improves both the standard accuracy and robustness. We think the reason lies in two aspects. First, compared with locally self-attention, convolutional FFN will not damage the long-term dependencies modeling ability of ViTs. The merit of ViTs can be inherited. Second, original FFN only encodes single token representation, while convolutional FFN encodes both the current token and its neighbors. Such information exchange within a local region makes ViTs more robust.

\subsection{Classification Head}
\textbf{F7: } \textbf{Is the classification token ($\mathtt{CLS}$) important for ViTs? The answer is not, and replacing $\mathtt{CLS}$ with global average pooling on output tokens even improves the robustness. } CNNs adopt a global average pooling layer before the classifier to integrate visual features at different spatial locations. This practice also inherently takes advantage of the translation invariance of the image. However, ViTs use an additional classification token ($\mathtt{CLS}$) to perform classification, are not translation-invariant. To get over this shortcoming, CPVT~\cite{chu2021we} and LeViT~\cite{graham2021levit} remove the $\mathtt{CLS}$ token and replace it by average pooling along with the last layer sequential output of the Transformer. We compare models trained with and without $\mathtt{CLS}$ token in Table~\ref{tab:analysis_other}. The result shows the adversarial robustness can be greatly improved by removing $\mathtt{CLS}$ token. Also we find removing $\mathtt{CLS}$ token has slight help for the standard accuracy, which can be benefited from the desired translation-invariance.

\subsection{Combination of Robust Components}
\label{sec:2.5}
In the above, we separately analyze the effect of each designed component in the ViTs. To make use of these findings, we combine the selected useful components, listed in follows: 1) Extract low-level feature of patches using a convolutional stem; 2) Adopt the multi-stage design of ViTs and avoid blocks with larger spatial resolution; 3) Choose a suitable number of heads; 4) Use convolution in FFN; 5) Replace $\mathtt{CLS}$ token with token feature pooling. As we find the effects of the above modifications are superimposed, we adopt all of these robust components into ViTs, the resultant model is called Robust Vision Transformer (RVT). RVT has achieved the new state-of-the-art robustness compared to other ViT variants. To further improve the performance, we propose two novel techniques, position-aware attention scaling and patch-wise data augmentation, to train our RVT. Both of them are also applicable to other ViT models.

\section{Position-Aware Attention Scaling}
In this section, we introduce our proposed position encoding mechanism called Position-Aware Attention Scaling (PAAS), which modifies the rescaling operation in the dot product attention to a more generalized version. To start with, we illustrate the scaled dot-product attention in transformer firstly. And then the modification of PAAS will be explained. 

\paragraph{Scaled Dot-product Attention.} Scaled dot-product attention is a key component in Multi-Head Self Attention layer (MHSA) of Transformer. MHSA first generates set of queries $Q\in \mathbb{R}^{N \times d}$, keys $K\in \mathbb{R}^{N \times d}$, values $V\in \mathbb{R}^{N \times d}$ with the corresponding projection. Then the query vector $q\in \mathbb{R}^{d}$ is matched against the each key vector in $K$. The output is the weighted sum of a set of $N$ value vectors $v$ based on the matching score. This process is called scaled dot-product attention:
\begin{equation}
\label{eq.sdpatt}
    \text{Attention}(Q, K, V) = \text{Softmax}(QK^T/\sqrt{d})V
\end{equation}

For preventing extremly small gradients and stabilizing the training process, each element in $QK^T$ multiplies by a constant $\frac{1}{\sqrt{d}}$ to be rescaled into a standard range. 

\paragraph{Position-Aware Attention Scaling.} In this work, a more effective position-aware attention scaling method is proposed. To make the original rescaling process of dot-product attention position-aware, we define a learnable position importance matrix $W_{p} \in \mathbb{R}^{N\times N}$, which presents the importance of each pair of $q$-$k$. The oringinal scaled dot-product attention is modified as follows:
\begin{equation}
\label{eq.sdpatt2}
    \text{Attention}(Q, K, V) = \text{Softmax}(QK^T \odot (W_{p}/\sqrt{d}))V
\end{equation}

\begin{figure}
\centering
\includegraphics[width=0.45\textwidth]{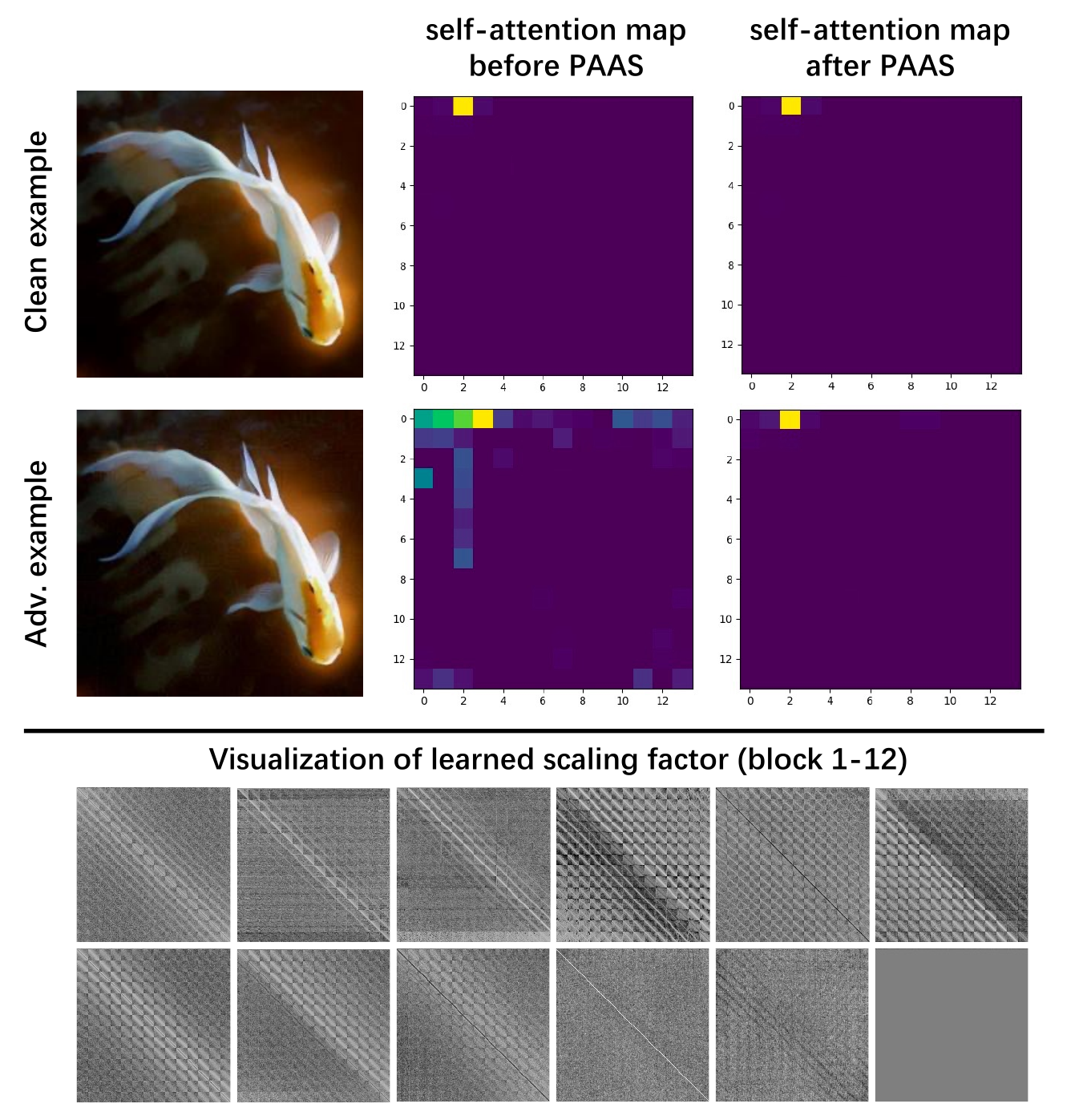}
\caption{\textbf{Top: }visualization of self-attention before and after the position-aware attention scaling. \textbf{Bottom: }visualization of learned scaling factor by our PAAS. }
\label{fig:robust_PAAS}
\vspace{-5mm}
\end{figure}

where $\odot$ is the element-wise product. As $W_{p}$ is input independent and only determined by the position of each $q$, $k$ in the sequence, our position-aware attention scaling can also serve as a position representation. Thus, we replace the traditional position embedding with our PAAS in RVT. After that the overall self-attention can be decoupled into two parts: the $QK^T$ term presents the content-based attention, and $W_{p}/\sqrt{d}$ term acts as the position-based attention. This untied design offers more expressiveness by removing the mixed and noisy correlations~\cite{ke2020rethinking}.

\paragraph{Robustness of PAAS.} As mentioned in section~\ref{sec:2.2}, most existing position embeddings have no contribution to the model robustness, and some of them even do a negative effect. Differently, our proposed PAAS can improve the model robustness effectively. This superior property relies on the position importance matrix $W_{p}$, which acts as a soft attention mask on each position pair of $q$-$k$. As shown in Figure~\ref{fig:robust_PAAS}, we visualize the attention map of 3\emph{th} query patch in 3\emph{th} transformer block. Without PAAS, an adversarial input can make some unrelated regions activated and produce a noisy self-attention map. To filter out these noises, PAAS suppresses the redundant positions irrelevant for classification in self-attention map, by a learned small multiplier in $W_{p}$. Finally only the regions important for classification are activated. We experimentally validate that PAAS can provide certain defense power against some white-box adversaries, e.g., FGSM~\cite{goodfellow2014explaining}. Not limited to adversarial attack, it also helps to the corruption and out-of-distribution generalization. Details can be referred to section~\ref{sec:5.3}.

\section{Patch-Wise Augmentation}
Image augmentation is a strategy especially important for ViTs since a biggest shortcoming of ViTs is the worse generalization ability when trained on relatively small-size datasets, while this shortcoming can be remedied by sufficient data augmentation~\cite{touvron2020training}. On the other hand, a rich data augmentation also helps with robustness and generalization, which has been verified in previous works~\cite{hendrycks2019augmix}. For improving the diversity of the augmented training data, we propose the patch-wise data augmentation strategy for ViTs, which imposes diverse augmentation on each input image patches at training time. Our motivation comes from the difference of ViTs and CNNs that ViTs not only extract intra-patch features but also concern the inter-patch relations. We think the traditional augmentation which randomly transforms the whole image could provide enough intra-patch augmentation. However, it lacks the diversity on inter-patch augmentation, as all of patches have the same transformation at one time. To impose more inter-patch diversity, we retain the original image-level augmentation, and then add 
the following patch-level augmentation on each image patch. For simplicity, only three basic image transformations are considered for patch-level augmentation: \emph{random resized crop}, \emph{random horizontal flip} and \emph{random gaussian noise}.

\paragraph{Robustness of Patch-Wise Augmentation.} Same with the augmentations like MixUp~\cite{zhang2017mixup}, AugMix~\cite{hendrycks2019augmix}, RandAugment~\cite{cubuk2020randaugment}, patch-wise augmentation also benefit the model robustness. It effects on the phases after conventional image-level augmentations, and provides the meaningful augmentation on patch sequence input. Different from RandAugment, which adopts augmentations conflicting with ImageNet-C, we only use simple image transform for patch-wise augmentation. It confirms that the most part of robustness improvement is derived from the strategy itself but not the used augmentation.
A significant advantage of patch-wise augmentation is that it can be in common use across different ViT models and bring more than 1\% and 5\% improvement on standard and robust accuracy. Details can be referred to section~\ref{sec:5.3}.

\section{Experiments}
\subsection{Experimental Settings}
\label{sec:6.1}
\textbf{Implementation Details.} All of our experiments are performed on the NVIDIA 2080Ti GPUs. We implement RVT in three sizes named by RVT-Ti, RVT-S, RVT-B respectively. All of them adopt the best settings investigated in section~\ref{sec:2}. For RVT$^{*}$, we add PAAS on multiple transformer blocks. The patch-wise augmentation uses the combination of base augmentation introduced in section~\ref{sec:5.4}. Other training hyperparameters are same with DeiT~\cite{touvron2020training}.

\textbf{Evaluation Benchmarks.} We adopt the ImageNet-1K~\cite{deng2009imagenet} dataset for training and standard performance evaluation. No other large-scale dataset is needed for pre-training. For robustness evaluation, we test our RVT in three aspect: 1) for adversarial robustness, we test the adversarial examples generated by white-box attack algorithms FGSM~\cite{goodfellow2014explaining} and PGD~\cite{madry2017towards} on ImageNet-1K validation set. ImageNet-A~\cite{hendrycks2019natural} is used for evaluating the model under natural adversarial example. 2) for common corruption robustness, we adopt ImageNet-C~\cite{hendrycks2019benchmarking} which consists of 15 types of algorithmically generated
corruptions with five levels of severity. 3) for out-of-distribution robustness, we evaluate on ImageNet-R~\cite{hendrycks2020many} and ImageNet-Sketch~\cite{wang2019learning}. They contain images with naturally occurring distribution changes. The difference is that ImageNet-Sketch only contains sketch images, which can be used for testing the classification ability when texture or color information is missing.

\subsection{Standard Performance Evaluation}
For standard performance evaluation, we compare our method with state-of-the-art classification methods including Transformer-based models and representative CNN-based models in Table~\ref{tab:sota_main}. Compared to CNNs-based models, RVT has surpassed most of CNN architectures with fewer parameters and FLOPs. RVT-Ti$^{*}$ achieves 79.2\% Top-1 accuracy on ImageNet-1K validation set, which
is competitive with currently popular ResNet and RegNet series, but only has 1.3G FLOPs and 10.9M parameters (around 60\% smaller than CNNs). With the same computation cost, RVT-S$^{*}$ obtains 81.9\% test accuracy, 2.9\% higher than ResNet-50. This result is closed to EfficientNet-B4, however EfficientNet-B4 requires larger 380$\times$380 input size and has much lower throughput.

\begin{table*}
    \centering
    \caption{The performance of RVT and several SOTA CNNs and Transformers on ImageNet and six robustness benchmarks. RVT$^{*}$ represents the RVT model but trained with our proposed PAAS and patch-wise augmentation. Except for different architectures, we also compare some methods such as AugMix, which aims at improving the model robustness based on ResNet-50.}
    \vspace{-2mm}
    \resizebox{1.8\columnwidth}{!}{
        \begin{tabular}{c|c|cc|cc|cccccc}
        \toprule
        \multirow{2}{*}{Group} &\multirow{2}{*}{Model} & FLOPs & Params & \multicolumn{2}{c|}{ImageNet} & \multicolumn{6}{c}{Robustness Benchmarks} \\
        & & (G) & (M) & Top-1 & Top-5 & FGSM & PGD & IN-C ($\downarrow$) & IN-A & IN-R & IN-SK\\
        \midrule
        \multirow{12}{*}{CNNs}&ResNet-50~\cite{DBLP:conf/cvpr/HeZRS16} & 4.1 & 25.6 & 76.1 & 86.0 & 12.2 & 0.9 & 76.7 & 0.0 & 36.1 & 24.1 \\
        &ResNet-50$^{*}$~\cite{DBLP:conf/cvpr/HeZRS16} & 4.1 & 25.6 & 79.0 & 94.4 & 36.3 & 12.5 & 65.5 & 5.9 & 42.5 & 31.5 \\
        &Inception v3~\cite{szegedy2016rethinking} & 5.7 & 27.2 & 77.4 & 93.4 & 22.5 & 3.1 & 80.6 & 10.0 & 38.9 & 27.6 \\
        &RegNetY-4GF~\cite{radosavovic2020designing} & 4.0 & 20.6 & 79.2 & 94.7 & 15.4 & 2.4 & 68.7 & 8.9 & 38.8 & 25.9 \\
        &EfficientNet-B4~\cite{tan2019efficientnet} & 4.4 & 19.3 & 83.0 & 96.3 & 44.6 & 18.5 & 71.1 & 26.3 & 47.1 & 34.1  \\
        &ResNeXt50-32x4d~\cite{xie2017aggregated} & 4.3 & 25.0 & 79.8 & 94.6 & 34.7 & 13.5 & 64.7 & 10.7 & 41.5 & 29.3 \\
        \cmidrule{2-12}
        &DeepAugment~\cite{hendrycks2020many} & 4.1 & 25.6 & 75.8 & 92.7 & 27.1 & 9.5 & 53.6 & 3.9 & 46.7 & 32.6 \\
        &ANT~\cite{rusak2020simple} & 4.1 & 25.6 & 76.1 & 93.0 & 17.8 & 3.1 & 63.0 & 1.1 & 39.0 & 26.3 \\
        &AugMix~\cite{hendrycks2019augmix} & 4.1 & 25.6 & 77.5 & 93.7 & 20.2 & 3.8 & 65.3 & 3.8 & 41.0 & 28.5\\
        &Anti-Aliased CNN~\cite{zhang2019making} & 4.2 & 25.6 & 79.3 & 94.6 & 32.9 & 13.5 & 68.1 & 8.2 & 41.1 & 29.6 \\
        &Debiased CNN~\cite{li2020shape} & 4.1 & 25.6 & 76.9 & 93.4 & 20.4 & 5.5 & 67.5 & 3.5 & 40.8 & 28.4 \\
        \midrule
        \multirow{24}{*}{Transformers}& DeiT-Ti~\cite{touvron2020training} & 1.3 & 5.7 & 72.2 & 91.1 & 22.3 & 6.2 & 71.1 & 7.3 & 32.6 & 20.2 \\
        & ConViT-Ti~\cite{d2021convit} & 1.4 & 5.7 & 73.3 & 91.8 & 24.7 & 7.5 & 68.4 & 8.9 & 35.2 & 22.4 \\
        & PiT-Ti~\cite{heo2021rethinking} & 0.7 & 4.9 & 72.9 & 91.3 & 20.4 & 5.1 & 69.1 & 6.2 & 34.6 & 21.6 \\
        & PVT-Tiny~\cite{wang2021pyramid} & 1.9 & 13.2 & 75.0 & 92.5 & 10.0 & 0.5 & 79.6 & 7.9 & 33.9 & 21.5 \\
        & RVT-Ti & 1.3 & 8.6 & 78.4 & 94.2 & 34.8 & 11.7 & 58.2 & 13.3 & 43.7 & 30.0 \\
        & \textbf{RVT-Ti$^{*}$} & 1.3 & 10.9 & \textbf{79.2} & \textbf{94.7} & \textbf{42.7} & \textbf{18.9} & \textbf{57.0} & \textbf{14.4} & \textbf{43.9} & \textbf{30.4} \\
         \cmidrule{2-12}
        & DeiT-S~\cite{touvron2020training} & 4.6 & 22.1 & 79.9 & 95.0 & 40.7 & 16.7 & 54.6 & 18.9 & 42.2 & 29.4 \\
        & ConViT-S~\cite{d2021convit} & 5.4 & 27.8 & 81.5 & 95.8 & 41.0 & 17.2 & 49.8 & 24.5 & 45.4 & 33.1 \\
        & Swin-T~\cite{liu2021swin} & 4.5 & 28.3 & 81.2 & 95.5 & 33.7 & 7.3 & 62.0 & 21.6 & 41.3 & 29.1 \\
        & PVT-Small~\cite{wang2021pyramid} & 3.8 & 24.5 & 79.9 & 95.0 & 26.6 & 3.1 & 66.9 & 18.0 & 40.1 & 27.2 \\
        & PiT-S~\cite{heo2021rethinking} & 2.9 & 23.5 & 80.9 & 95.3 & 41.0 & 16.5 & 52.5 & 21.7 & 43.6 & 30.8 \\
        & TNT-S~\cite{han2021transformer} & 5.2 & 23.8 & 81.5 & 95.7 & 33.2 & 4.2 & 53.1 & 24.7 & 43.8 & 31.6 \\
        & T2T-ViT\_t-14~\cite{yuan2021tokens} & 6.1 & 21.5 & 81.7 & \textbf{95.9} & 40.9 & 11.4 & 53.2 & 23.9 & 45.0 & 32.5 \\
        & RVT-S & 4.7 & 22.1 & 81.7 & 95.7 & 51.3 & 26.2 & 50.1 & 24.1 & 46.9 & \textbf{35.0} \\
        & \textbf{RVT-S$^{*}$} & 4.7 & 23.3 & \textbf{81.9} & 95.8 & \textbf{51.8} & \textbf{28.2} & \textbf{49.4} & \textbf{25.7} & \textbf{47.7} & 34.7  \\
        \cmidrule{2-12}
        & DeiT-B~\cite{touvron2020training} & 17.6 & 86.6 & 82.0 & 95.7 & 46.4 & 21.3 & 48.5 & 27.4 & 44.9 & 32.4 \\
        & ConViT-B~\cite{d2021convit} & 17.7 & 86.5 & 82.4 & 96.0 & 45.4 & 20.8 & 46.9 & 29.0 & 48.4 & 35.7 \\
        & Swin-B~\cite{liu2021swin} & 15.4 & 87.8 & \textbf{83.4} & 96.4 & 49.2 & 21.3 & 54.4 & \textbf{35.8} & 46.6 & 32.4 \\
        & PVT-Large~\cite{wang2021pyramid} & 9.8 & 61.4 & 81.7 & 95.9 & 33.1 & 7.3 & 59.8 & 26.6 & 42.7 & 30.2\\
        & PiT-B~\cite{heo2021rethinking} & 12.5 & 73.8 & 82.4 & 95.7 & 49.3 & 23.7 & 48.2 & 33.9 & 43.7 & 32.3 \\
        & T2T-ViT\_t-24~\cite{yuan2021tokens} & 15.0 & 64.1 & 82.6 & 96.1 & 46.7 & 17.5 & 48.0 & 28.9 & 47.9 & 35.4 \\
        & RVT-B & 17.7 & 86.2 & 82.5 & 96.0 & 52.3 & 27.4 & 47.3  & 27.7 & 48.2 & 35.8 \\
        & \textbf{RVT-B$^{*}$} & 17.7 & 91.8 & 82.7 & \textbf{96.5} & \textbf{53.0} & \textbf{29.9} & \textbf{46.8} & 28.5 & \textbf{48.7} & \textbf{36.0}    \\

        \bottomrule
    \end{tabular}
    }
    \vspace{-6mm}
    \label{tab:sota_main}
\end{table*}

Compared to Transformer-based models, our RVT also achieves the comparable standard accuracy. We find just combining the robust components can make RVT-Ti get 78.4\% Top-1 accuracy and surpass the existing state-of-the-art on ViTs with tiny version. By adopting our newly proposed position-aware attention scaling and patch-wise data augmentation, RVT-Ti$^{*}$ can further 
improve 0.8\% on RVT-Ti with little additional computation cost. For other scales of the model, RVT-S$^{*}$ and RVT-B$^{*}$ also achieve a good promotion compared with DeiT-S and DeiT-B. Although the improvement becomes smaller with the increase of model capacity, we think the advance of our model is still obvious as it strengthen the model ability in various views such as robustness and out-of-domain generalization.

\subsection{Robustness Evaluation}
\label{sec:5.3}
We employ a series of benchmarks to evaluate the model robustness on different aspects. Among them, ImageNet-C (IN-C) calculates the mean corruption error (mCE) as metric. The smaller mCE means the more robust of the model under corruptions. All other benchmarks use Top-1 accuracy on test data if no special illustration. The results are reported in Table~\ref{tab:sota_main}.

\textbf{Adversarial Robustness.} For evaluating the adversarial robustness, we adopt single-step attack algorithm FGSM~\cite{goodfellow2014explaining} and multi-step attack algorithm PGD~\cite{madry2017towards} with steps $t=5$, step size $\alpha=0.5$. Both attackers perturb the input image with max magnitude $\epsilon=1$. Table~\ref{tab:sota_main} suggests that the adversarial robustness has a strong correlation with the design of model architecture. With similar model scale and FLOPs, most Transformer-based models have higher robust accuracy than CNNs under adversarial attacks. This conclusion is also consistent with \cite{shao2021adversarial}. Some modifications on ViTs or CNNs will also weaken or strengthen the adversarial robustness. For example, Swin-T~\cite{liu2021swin} introduces window self-attention for reducing the computation cost but damaging the adversarial robustness, and EfficientNet-B4\cite{tan2019efficientnet} uses smooth activation functions which is helpful with adversarial robustness.

We summarize the robust design experiences of ViTs in this work. The resultant RVT model achieves superior performance on both FGSM and PGD attackers. In detail, RVT-Ti and RVT-S get over 10\% improvement on FGSM, compared with the previous ViT variants. This advance is further expanded by our PAAS and patch-wise augmentation. Adversarial robustness seems unrelated with standard performance. Although models like Swin-T, TNT-S get higher standard accuracy than DeiT-S, their adversarially robust accuracy is well below the baseline. However, our RVT model can achieve the best trade-off between standard performance and adversarial robustness.

\textbf{Common Corruption Robustness.}
To metric the model degradation on common image corruptions, we present the mCE on ImageNet-C (IN-C) in Table~\ref{tab:sota_main}. We also list some methods from ImageNet-C Leaderboard, which are built based on ResNet-50. Our RVT-S$^{*}$ gets 49.4 mCE, which has 4.2 improvement on top-1 method DeepAugment~\cite{hendrycks2020many} in the leaderboard, and bulids the new state-of-the-art. The result also indicates that Transformer-based models have a natural advantage in dealing with image corruptions. Attributed to its ability of long-range dependencies modeling, ViTs are easier to learn the shape-bias features. Note that in this work we are not considering RandAugment. As a training augmentation of ViTs, RandAugment adopts conflicted augmentation with ImageNet-C and may cause the unfairness of the comparison proposed by~\cite{bai2021transformers}.

\textbf{Out-of-distribution Robustness.} We test the generalization ability of RVT on out-of-distribution data by reporting the top@1 accuracy on ImageNet-R (IN-R) and ImageNet-Sketch (IN-SK) in Table~\ref{tab:sota_main}. Our RVT and RVT$^{*}$ also beat other ViT models on out-of-distribution generalization. As the superiority of Transformer-based models on capturing shape-bias features mentioned above, our RVT-S also surpasses most CNN and ViT models and get 35.0\% and 46.9\% test accuracy on ImageNet-Sketch and ImageNet-R, buliding the new state-of-the-art. 
\begin{minipage}{\textwidth}
\begin{minipage}[t]{0.216\textwidth}
\makeatletter\def\@captype{table}
\small
    \centering
     \tablestyle{5pt}{1.05}
\begin{tabular}{c|c|c|c}
\toprule

\multirow{2}{*}{Layers} & Pos. & \multirow{2}{*}{Acc} & Rob. \\
& Emb. &  & Acc \\
\midrule
 \multirow{2}{*}{0-1} & Ori. & 78.2 & 34.1 \\
& \underline{Ours} & 78.4 & 34.3 \\
 \multirow{2}{*}{0-5} & Ori. & 78.4 & 34.6 \\
& \underline{Ours} & 78.6 & 35.2 \\
 \multirow{2}{*}{0-10} & Ori. & 78.4 & 34.8 \\
& \underline{Ours} & \textbf{78.6} &  \textbf{35.3} \\
\bottomrule
\end{tabular}

\caption{Comparison of single and multiple block PAAS. Ori. stands for the learned absolute position embedding in original ViTs.}
\label{ablate:1}
\end{minipage}
\begin{minipage}[t]{0.255\textwidth}
\makeatletter\def\@captype{table}
\small
    \centering
     \tablestyle{5pt}{1.05}
\begin{tabular}{c|c|c|c|c}
\toprule

\multicolumn{3}{c|}{Augmentations} & \multirow{2}{*}{Acc} & \multirow{2}{*}{Rob. Acc} \\
 RC & GN & HF & & \\
\midrule
\cmark & &  & 78.9 & 41.5 \\
 & \cmark &  & 79.0 & \textbf{42.0} \\
 & & \cmark & 79.1  & 41.3 \\
\cmark &  & \cmark & 78.8 & 41.3 \\
 & \cmark & \cmark & 79.0 & 41.9 \\
\cmark & \cmark & \cmark & \textbf{79.2} & 41.7 \\
\bottomrule
\end{tabular}

\caption{Ablation experiments on patch-wise augmentation. RC, GN, HF represent \emph{random resized crop}, \emph{random gaussian noise} and \emph{random horizontal flip} respectively.}
\label{ablate:2}
\vspace{-10pt}
\end{minipage}
\end{minipage}

\subsection{Ablation Studies}
\label{sec:5.4}
we conduct ablation studies on the proposed components of PAAS and patch-wise augmentation in this section. Other modifications of RVT are not involved since they have been analyzed in section~\ref{sec:2}. All of our ablation experiments are based on the RVT-Ti model on ImageNet.



\textbf{Single layer PAAS vs. Multiple layer PAAS.} We evaluate whether using PAAS on multiple transformer blocks can benefit the performance or robustness. The result is suggested in Table~\ref{ablate:1}. Learned absolute position embedding in original ViT model is adopted for comparison. With more transformer blocks using PAAS, the standard and robust accuracy gain greater enhancement. After applying PAAS on 5 blocks, the benefit of PAAS gets saturated. There will be the same trend if we replace PAAS with the original position embedding. But the original position embedding is not performed as good as our PAAS on both standard and robust accuracy.

\textbf{Different types of basic augmentation.} Due to the limited training resources, we only test three basic image augmentations: \emph{random resized crop}, \emph{random horizontal flip} and \emph{random gaussian noise}. For random resized crop, we crop the patch according to the scale sampled from [0.85, 1.0], then resize it to original size with aspect ratio unchanged. We set the mean and standard deviation as 0 and 0.01 for random gaussian noise. For each transformation, we set the applying probability $p=0.1$. Other hyper-parameters are consistent with the implementation in Kornia~\cite{eriba2019kornia}. As shown in Table~\ref{ablate:2}, we can see both three augmentations are beneficial of 
standard and robust accuracy. Among them, random gaussian noise is the better choice as it helps for more robustness improvement.

\textbf{Combination of basic augmentations.} We further evaluate the combination of basic patch-wise augmentations. For traditional image augmentation, combining multiple basic transformation~\cite{cubuk2020randaugment} can largely improve the standard accuracy. Differently, as shown in Table~\ref{ablate:2}, the benefit is marginal for combining basic patch-wise augmentations, but combination of three is still better than using only single augmentation. In this paper, we adopt the combination of all basic augmentations.

\textbf{Effect on other ViT architectures.}
For showing the effectiveness of our proposed position-aware attention scaling and patch-wise augmentation, we apply them to train other ViT models. DeiT-Ti, ConViT-Ti and PiT-Ti are adopted as the base model. The experimental results are shown in Table~\ref{ablate:3}, with combining the proposed techniques into these base models, all the augmented models achieve significant improvement. Specifically, all the improved models yield more than 1\% and 5\% promotion on standard and robust accuracy on average.

\begin{table}[h!]
	\vspace{-5pt}
    \small
    \centering
     \tablestyle{5pt}{1.05}
\begin{tabular}{c|c|c|c|c|c}
\toprule

Vanilla & \multirow{2}{*}{Acc} & \multirow{2}{*}{Rob. Acc} & Improved & \multirow{2}{*}{Acc} & \multirow{2}{*}{Rob. Acc} \\
models &  &  & models & &  \\
\midrule
DeiT-Ti & 72.2 & 22.3 & DeiT-Ti$^{*}$ & \textbf{74.4} & \textbf{29.9} \\
ConViT-Ti & 73.3 & 24.7 & ConViT-Ti$^{*}$ & \textbf{74.4} & \textbf{30.7} \\
PiT-Ti & 72.9 & 20.4 & PiT-Ti$^{*}$ & \textbf{74.3} & \textbf{27.7} \\
\bottomrule
\end{tabular}

\caption{Effect of our proposed PAAS and patch-wise augmentation on other ViT architectures.}
\label{ablate:3}
\vspace{-10pt}
\end{table}

\section{Conclusion}

We systematically study the robustness of key components in ViTs, and propose Robust Vision Transformer (RVT) by alternating the modifications which would damage the robustness. Furthermore, we have devised a novel patch-wise augmentation which adds rich affinity and diversity to training data. Considering the lack of spatial information correlation in scaled dot-product attention, we present position-aware attention scaling (PAAS) method to further boost the RVT. Experiments show that our RVT achieves outstanding performance consistently on ImageNet and six robustness benchmarks. Under the exhaustive trade-offs between FLOPs, standard and robust accuracy, extensive experiment results validate the significance of our RVT-Ti and RVT-S.

{\small
\bibliographystyle{ieee_fullname}
\bibliography{egbib}
}

\clearpage

\appendix

\section*{Appendix}

\section{Additional Results of Robustness Analysis on Designed Components}

Here we will show the remaining results of robustness analysis in section~\ref{sec:3}. As the each component has been discussed detailly, we only give a summary of the results in appendix. We report the additional results of robustness analysis in Table~\ref{tab:appendix_posemb},~\ref{tab:appendix_head},~\ref{tab:appendix_stage},~\ref{tab:appendix_others},~\ref{tab:appendix_othervit} and ~\ref{tab:appendix_finetune22k} respectively, where each table presents the results of one or some components. The detailed architecture of models used in robustness analysis on stage distribution is shown in Table~\ref{tab:appendix_arch}. Although each robustness benchmark is consistent on the overall trend, we still find some special cases. For example, in Table~\ref{tab:appendix_stage}, the V6 version of stage distribution poorly performs on adversarial robustness, but achieves best results on IN-A and IN-R datasets, showing the superior generalization power. Another case is the token-to-token embedder in Table~\ref{tab:appendix_others}. Compared with original linear embedder, token-to-token embedder obtains better results on IN-C, IN-A, IN-R and IN-SK datasets. However, under PGD attacker, it only gets the robust accuracy of 4.7\%. The above phenomenon also indicates that using only several robustness benchmarks is biased and cannot get a comprehensive assessment result. Therefore, we advocate that the works about model robustness in future should consider multiple benchmarks. For validating the generality of the proposed techniques, we show the robustness evaluation results when trained on other ViT architectures and larger datasets (ImageNet-22k) in Table~\ref{tab:appendix_othervit} and ~\ref{tab:appendix_finetune22k}.

\begin{table}[h]
\footnotesize
    \centering

\begin{tabular}{c|c|c|c|c|c|c}
	\toprule
	Heads & 1 & 2 & 4 & 6 & 8 & 12  \\
	\midrule
	Acc & 69.0 & 71.7 & 73.1 & 73.4 & \textbf{73.9} & 73.5 \\
	FGSM & 17.6 & 21.4 & 22.8 & 24.6 & \textbf{25.2} & 24.7 \\ 
	PGD & 4.3 & 6.1 & 7.1 & 7.7 & \textbf{8.2} & 8.0 \\ 
	IN-C ($\downarrow$) & 79.5 & 72.9 & 69.0 & 68.5 & \textbf{67.7} & 68.2 \\ 
	IN-A & 5.1 & 6.9 & 8.2 & 8.3 & \textbf{8.9} & 8.4 \\ 
	IN-R & 28.1 & 32.9 & 33.9 & 34.1 & \textbf{34.2} & 33.7 \\ 
	IN-SK & 15.9 & 20.4 & 21.4 & 21.6 & \textbf{22.0} & 21.1 \\ 
	
	\bottomrule
\end{tabular}
\vspace{2mm}
    \caption{Additional results of robustness analysis on different head number.}
    \label{tab:appendix_head}
\end{table}

\begin{table*}[h]
\footnotesize
    \centering
\begin{tabular}{lr|c|c|c|c|c|c|c}
\toprule
&Position Embeddings & Acc & FGSM & PGD & IN-C ($\downarrow$) & IN-A & IN-R & IN-SK  \\
\midrule
&none   & 68.3 & 15.8 &3.6 & 82.4 & 5.2 &24.3 & 12.0\\ 
&learned absolute Pos. & 72.2 & \textbf{22.3} & \textbf{6.2} & \textbf{71.1} & 7.3 & \textbf{32.6} & \textbf{20.3}  \\
&sin-cos absolute Pos. & 72.0 & 21.9 & 5.9 & 71.9 & 7.0 & 31.4 & 20.2  \\ 
&learned relative Pos.\cite{shaw2018self}  & 71.8 & 22.3 & 6.1 &71.6 &\textbf{7.6} & 32.5 & 18.6 \\
&input-conditioned Pos.\cite{chu2021we} & \textbf{72.4} & 21.5 & 5.3 & 72.5 & 6.8 & 31.0 & 18.0 \\
\bottomrule
\end{tabular}
\vspace{2mm}
    \caption{Additional results of robustness analysis on different position encoding methods.}
    \label{tab:appendix_posemb}
\end{table*}

\begin{table*}[h]
\footnotesize
    \centering

\begin{tabular}{c|c|c|c|c|c|c|c|c|c|c}
  \toprule
  \multicolumn{3}{c|}{Patch Emb.} 
  & Local & Conv. & \multirow{2}{*}{$\mathtt{CLS}$} &  \multirow{2}{*}{PGD}&
		\multirow{2}{*}{IN-C ($\downarrow$)}&
		\multirow{2}{*}{IN-A}&
		\multirow{2}{*}{IN-R}&
		\multirow{2}{*}{IN-SK}\\
  Linear & Conv. & T2T & SA & FFN & & & & & \\
  \midrule
  \cmark & &  &  & & \cmark & 6.2 & 71.1 & 7.3 & 32.6 & 20.3 \\
  & \cmark & &  & & \cmark & 6.8 & 69.2 & 8.3 & 33.6 & 21.1\\
  &  & \cmark &  & & \cmark & 4.7 & 69.6 & 10.1 & 36.7 & 23.8\\
  \cmark &  & & \cmark  &  & \cmark & 9.0 & 76.9 & 4.8 & 28.7 & 16.6 \\
  \cmark &  & & & \cmark & \cmark & 12.7 & 65.0 & 8.4 & 39.0 & 31.9\\
  \cmark &  & &  &  &  & 12.0 & 70.0 & 7.4 & 32.5 & 20.2 \\
  \bottomrule
  \end{tabular}
  \vspace{2mm}
    \caption{Additional results of robustness analysis on different patch embeddings, locality of attention, convolutional FFN and the replacement of $\mathtt{CLS}$ token.}
    \label{tab:appendix_others}
\end{table*}

\begin{table*}[h]
\footnotesize
    \centering

\begin{tabular}{c|c|c|c|c|c|c|c|c}
        \toprule
		\multirow{2}{*}{Var.} &\multirow{2}{*}{{[{S$_{1}$}, {S$_{2}$}, {S$_{3}$}, {S$_{4}$}]}} &  \multirow{2}{*}{Acc} & \multirow{2}{*}{FGSM} & \multirow{2}{*}{PGD}&
		\multirow{2}{*}{IN-C ($\downarrow$)}&
		\multirow{2}{*}{IN-A}&
		\multirow{2}{*}{IN-R}&
		\multirow{2}{*}{IN-SK}\\
		& & & & & & & &\\
			\midrule
			V1 & [0, 0, 12, 0] & 72.2 & 22.3 & 6.2 & 71.1 & 7.3 & 32.6 & 20.3 \\ 
			\underline{V2} & \underline{[0, 0, 10, 2]} & 74.8 & \textbf{24.3} & \textbf{6.8} & \textbf{66.9} & 8.8 & 35.5 & 21.9\\ 
			V3 & [0, 2, 10, 0] & 73.8 & 22.0 & 5.1 & 76.4 & 8.2 & 33.6 & 21.1  \\ 
			V4 & [0, 2, 8, 2] & 76.4 &  22.3 & 4.5 & 71.5 & 10.3 & 36.8 & \textbf{23.9} \\ 
			V5 & [2, 2, 8, 0] & 73.4 & 17.0 & 2.3 & 76.8 & 9.0 & 33.2 & 20.7 \\ 
			V6 & [2, 2, 6, 2] & \textbf{76.4} & 17.5 & 1.9 & 71.6 & \textbf{11.2} & \textbf{36.8} & 23.1 \\ 
			\bottomrule
  \end{tabular}
  \vspace{2mm}
    \caption{Additional results of robustness analysis on stage distribution.}
    \label{tab:appendix_stage}
\end{table*}

\begin{table*}[h]
\footnotesize
    \centering
\begin{tabular}{lr|c|c|c|c|c|c|c}
\toprule
&Models & Acc & FGSM & PGD & IN-C ($\downarrow$) & IN-A & IN-R & IN-SK  \\
\midrule

&DeiT-Ti   & 72.2 & 22.3 &6.2 & 71.1 & 7.3 &32.6 & 20.2\\ 
&DeiT-Ti$^{*}$ & 74.4 & 29.9 & 9.1 & 67.9 & 8.1 & 34.9 & 23.1  \\
&ConViT-Ti & 73.3 & 24.7 & 7.5 & 68.4 & 8.9 & 35.2 & 22.4  \\ 
&ConViT-Ti$^{*}$ & 74.4 & 30.7 & 9.6 &65.6 &9.4 & 37.0 & 25.2 \\
&PiT-Ti & 72.9 & 20.4 & 5.1 & 69.1 & 6.2 & 34.6 & 21.6 \\
&PiT-Ti$^{*}$ & 74.3 & 27.7 & 7.9 & 66.7 & 7.1 & 36.6 & 24.0 \\
\\

&DeiT-S   & 79.9 & 40.7 &16.7 & 54.6 & 18.9 &42.2 & 29.4\\ 
&DeiT-S$^{*}$ & 80.6 & 42.3 & 18.8 & 53.1 & 20.5 & 43.5 & 31.3  \\
&ConViT-S & 81.5 & 41.0 & 17.2 & 49.8 & 24.5 & 45.4 & 33.1  \\ 
&ConViT-S$^{*}$ & 81.8 & 42.3 & 18.7 & 49.1 & 25.6 & 46.1 & 34.2 \\
&PiT-S & 80.9 & 41.0 & 16.5 & 52.5 & 21.7 & 43.6 & 30.8 \\
&PiT-S$^{*}$ & 81.4 & 42.2 & 18.3 & 51.4 & 23.3 & 44.6 & 32.3 \\
\bottomrule
\end{tabular}
\vspace{2mm}
    \caption{Additional results of position-aware attention scaling and patch-wise augmentation on other ViT architectures.}
    \label{tab:appendix_othervit}
\end{table*}

\begin{table*}[h]
\footnotesize
    \centering
\begin{tabular}{lr|c|c|c|c|c|c|c}
\toprule
&Models & Acc & FGSM & PGD & IN-C ($\downarrow$) & IN-A & IN-R & IN-SK  \\
\midrule

&DeiT-B   & 83.20 & 47.21 &24.89 & 45.50 & 38.01 &52.37 & \textbf{39.54}\\ 
&RVT-B & 83.57 & 53.67 & 30.45 & 44.26 & 41.00 & 49.67 & 35.01  \\
&RVT-B$^{*}$ & \textbf{83.80} & \textbf{55.40} & \textbf{33.86} & \textbf{42.99} & \textbf{42.27} & \textbf{52.63} & 38.43 \\

\bottomrule
\end{tabular}
\vspace{2mm}
    \caption{RVT pre-trained on ImageNet-22K and finetuned on ImageNet-1K.}
    \label{tab:appendix_finetune22k}
\end{table*}

\section{Feature Visualization}
\begin{figure}
    \centering
    \includegraphics[width=1.\linewidth]{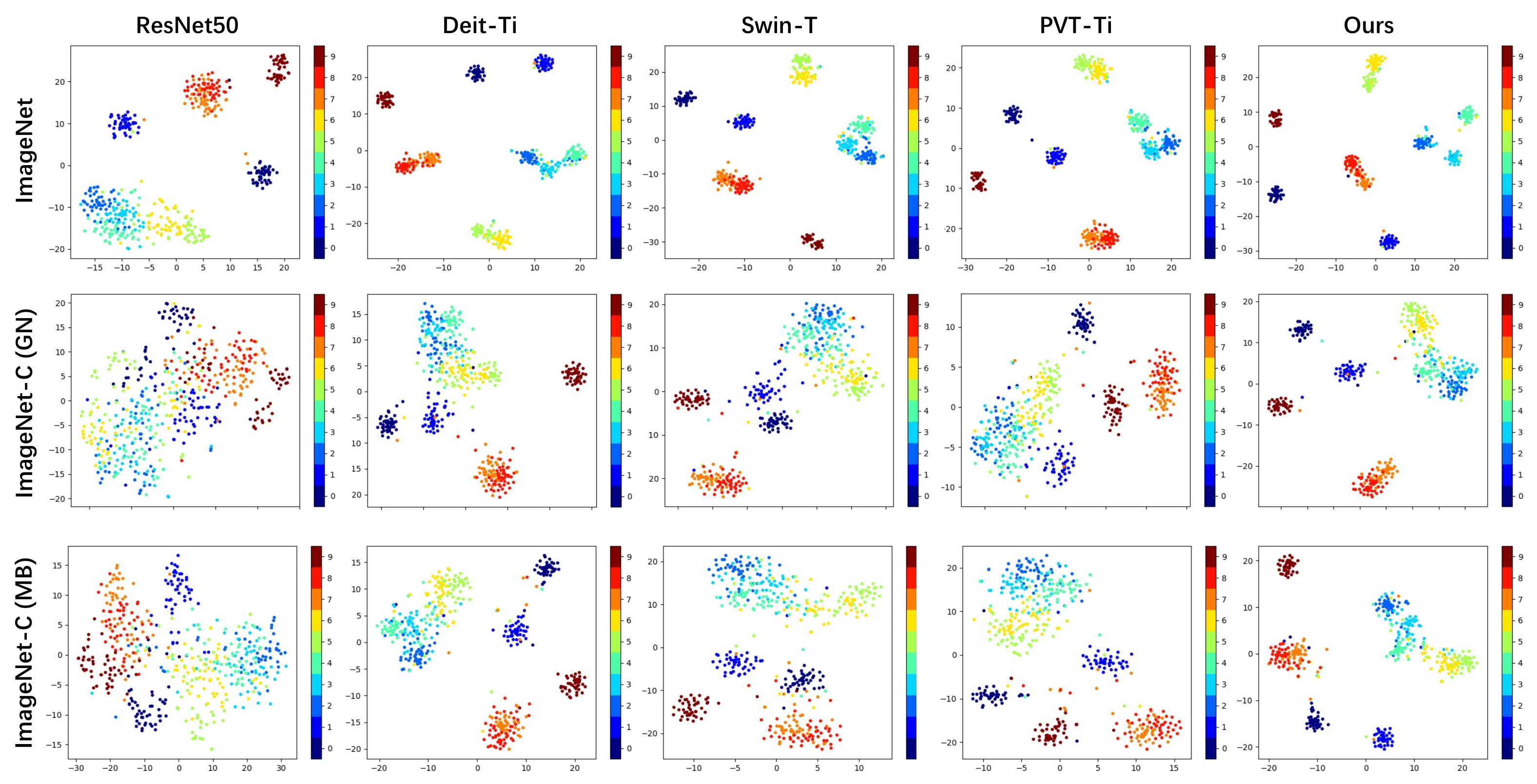}
    \vspace{-4mm}
    \caption{\textbf{t-SNE} visualization of features produced by different models.}
    \label{fig:tsne_vis}
\end{figure}

In general understanding, intra-class compactness and inter-class separability are crucial indicators to measure the effectiveness of a model to
produce discriminative and robust features. We use t-Distributed Stochastic Neighbor Embedding (t-SNE) to visualize the feature sets extracted by ResNet50, DeiT-Ti, Swin-T, PVT-Ti and our RVT respectively. The features are produced on validation set of ImageNet and ImageNet-C. We randomly selected 10 classes for better visualization. As shown in Figure~\ref{fig:tsne_vis}, features extracted by our RVT is the closest to the intra-class compactness and inter-class separability. It's confirmed from the side that our RVT does have the stronger robustness and classification performance. 

\begin{figure}
    \centering
    \includegraphics[width=1.\linewidth]{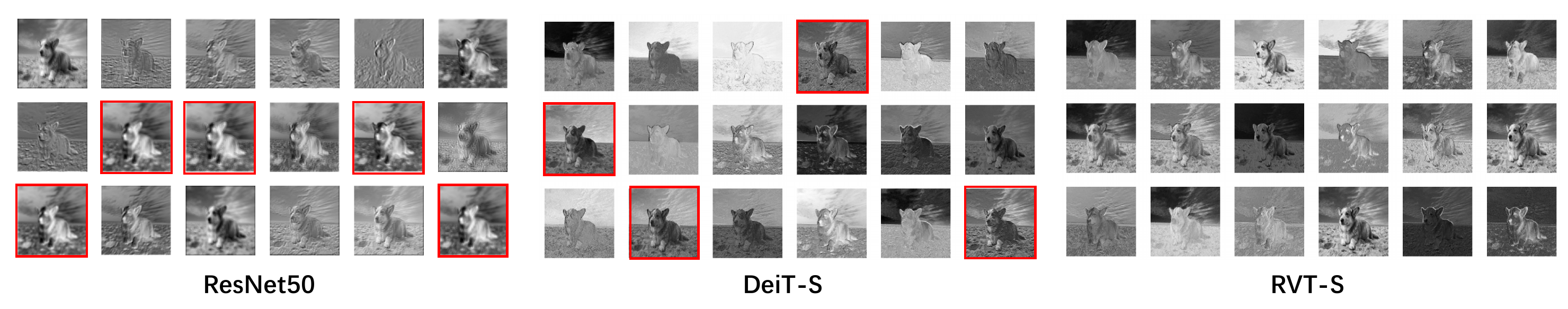}
    \vspace{-4mm}
    \caption{Feature visualization of ResNet50, DeiT-S and our proposed RVT-S trained on ImageNet. Red boxes highlight the feature maps with high similarity. }
    \label{fig:feature_vis}
\end{figure}

We also visualize the feature maps of ResNet50, DeiT-S and our proposed RVT-S in Figure~\ref{fig:feature_vis}. Visualized features are extracted on the 5th layer of the models. The result shows ResNet50 and DeiT-S contain a large part of redundant features, highlighted by red boxes. While our RVT-S reduces the redundancy and ensures the diversity of features, reflecting the stronger generalization ability.

\section{Loss Landscape Visualization}
\begin{figure}
    \centering
    \includegraphics[width=1.\linewidth]{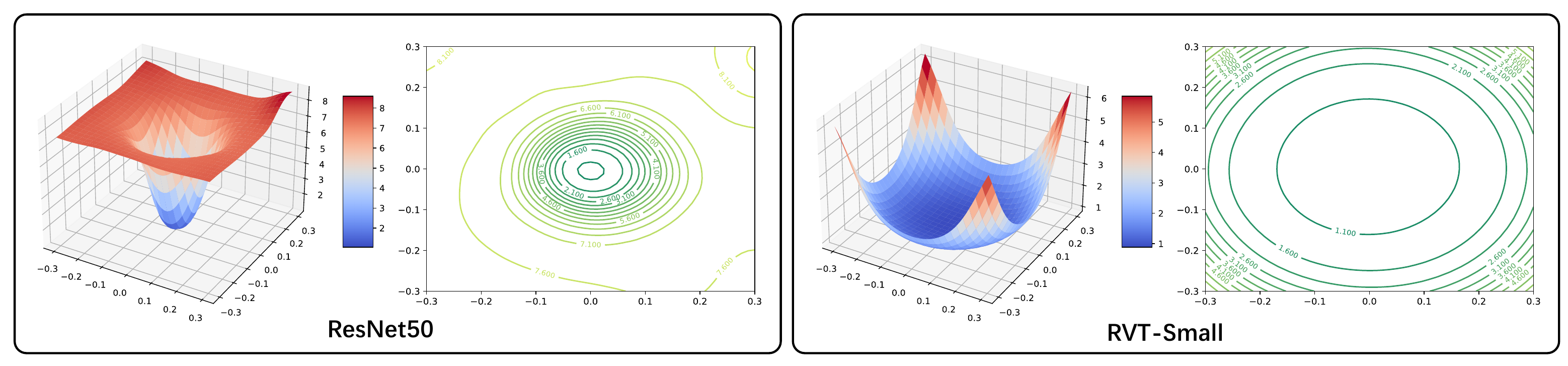}
    \vspace{-4mm}
    \caption{Loss landscape of ResNet50 and RVT-S.}
    \label{fig:loss_vis}
\end{figure}

Loss landscape geometry has a dramatic effect on generalization and trainability of the model. We visualize the loss surfaces of ResNet50 and our RVT-S in Figure~\ref{fig:loss_vis}. RVT-S has a flatter loss surfaces, which means the stability under input changes.

\begin{table*}[h]
\scriptsize
    \centering
\begin{tabular}{c|c|c|c|c|c}
	 & Output Size & Layer Name & DeiT-Ti (V1) & V4 & V5 \\
	\toprule
	\multirow{2}{*}[-2.5ex]{Stage 1} & \multirow{2}{*}[-2.5ex]{\scalebox{1.3}{$\frac{H}{4}\times \frac{W}{4}$}} & Patch Embedding & $C_{1}=192$ & $C_{1}=96$ & $C_{1}=48$ \\
	\cline{3-6}
	& & \tabincell{c}{Transformer\\Encoder} & - & - &
	$\begin{bmatrix}
	\begin{array}{l}
	H_1=48 \\
	N_1=1 \\
	C_1=48 \\
	\end{array}
	\end{bmatrix} \times 2$ \\
	\hline
	\multirow{2}{*}[-2.5ex]{Stage 2} & \multirow{2}{*}[-2.5ex]{\scalebox{1.3}{$\frac{H}{8}\times \frac{W}{8}$}} & Pooling Layer & - & - & $k=2\times2$ \\
	\cline{3-6}
	& & \tabincell{c}{Transformer\\Encoder} & - &
	$\begin{bmatrix}
	\begin{array}{l}
	H_2=48 \\
	N_2=2 \\
	C_2=96 \\
	\end{array}
	\end{bmatrix} \times 2$ &
	$\begin{bmatrix}
	\begin{array}{l}
	H_2=48 \\
	N_2=2 \\
	C_2=96 \\
	\end{array}
	\end{bmatrix} \times 2$ \\
	\hline
	\multirow{2}{*}[-2.5ex]{Stage 3} & \multirow{2}{*}[-2.5ex]{\scalebox{1.3}{$\frac{H}{16}\times \frac{W}{16}$}} & Pooling Layer  & - & $k=2\times2$ & $k=2\times2$ \\
	\cline{3-6}
	& & \tabincell{c}{Transformer\\Encoder} &
	$\begin{bmatrix}
	\begin{array}{l}
	H_2=64 \\
	N_2=3 \\
	C_2=192 \\
	\end{array}
	\end{bmatrix} \times 12$ &
	$\begin{bmatrix}
	\begin{array}{l}
	H_3=64 \\
	N_3=3 \\
	C_3=192 \\
	\end{array}
	\end{bmatrix} \times 8$ &
	$\begin{bmatrix}
	\begin{array}{l}
	H_3=64 \\
	N_3=3 \\
	C_3=192 \\
	\end{array}
	\end{bmatrix} \times 6$ \\
	\hline
	\multirow{2}{*}[-2.5ex]{Stage 4} &  \multirow{2}{*}[-2.5ex]{\scalebox{1.3}{$\frac{H}{32}\times \frac{W}{32}$}} & Pooling Layer & - & $k=2\times2$ & $k=2\times2$ \\
	\cline{3-6}
	& & \tabincell{c}{Transformer\\Encoder} & - &
	$\begin{bmatrix}
	\begin{array}{l}
	H_3=64 \\
	N_3=6 \\
	C_3=384 \\
	\end{array}
	\end{bmatrix} \times 2$ & $\begin{bmatrix}
	\begin{array}{l}
	H_4=64 \\
	N_4=6 \\
	C_4=384 \\
	\end{array}
	\end{bmatrix} \times 2$ \\
\end{tabular}
\vspace{2mm}
    \caption{Detailed architecture of models used in robustness analysis on stage distribution. $C$, $H$ and $N$ represent the total feature dimension, feature dimension of each head and head number respectively. Only V4 and V5 are listed as examples. The other versions of the model can be generalized by V4 and V5. }
    \label{tab:appendix_arch}
\end{table*}

\end{document}